\documentclass[10pt,twocolumn,letterpaper]{article}

\usepackage{cvpr}
\usepackage{times}
\usepackage{epsfig}
\usepackage{graphicx}
\usepackage{amsmath}
\usepackage{amssymb}
\usepackage{times}  
\usepackage{helvet}  
\usepackage{courier}  
\usepackage{url}  
\usepackage{graphicx}  
\usepackage{comment}
\usepackage{times}
\usepackage{graphicx}
\usepackage{amssymb}
\usepackage{graphicx}
\usepackage{verbatim}
\usepackage{mathrsfs}
\usepackage{mathtools}
\usepackage[english]{babel} 
\usepackage{mathrsfs}
\usepackage{array}
\newcolumntype{P}[1]{>{\centering\arraybackslash}p{#1}}
\usepackage{bm}
\newcommand{\Tref}[1]{Table~\ref{#1}}
\newcommand{\Eref}[1]{Equation~(\ref{#1})}
\newcommand{\Fref}[1]{Figure~\ref{#1}}
\newcommand{\Sref}[1]{Section~\ref{#1}}

\newcommand{\boxin}[1]{\textcolor{magenta}{{[Boxin: #1]}}}

\def\etal{\emph{ et al.}}
\def\ie{\emph{i.e.}}
\def\eg{\emph{e.g.}}
\def\x{{\mathbf x}}
\def\F{{\mathcal{F}}}
\def\eg{{\emph{e.g.}}}
\def\ie{{\emph{i.e.}}}
\def\etal{{\emph{et al.}}}
\def\I{{\mathbf{I}}}
\def\B{{\mathbf{B}}}
\def\R{{\mathbf{R}}}
\def\M{{\mathbf{M}}}

\def\G{{\mathbf{\mathcal{G}}}}

\def\E{{\mathbb{E}}}

\def\H{{\mathcal{H}}}
\def\W{{\mathbf{W}}}
\def\G{{\mathcal{G}}}

\def\mmu{\bm{\mu}}
\def\P{\mathbb{P}}
\def\ll{\mathcal{L}}
\def\Ee{{\mathbf{E}}}
\def\pphi{{\Phi}}
\def\ttheta{\bm{\theta}}

\usepackage[breaklinks=true,bookmarks=false]{hyperref}

\cvprfinalcopy 

\def\cvprPaperID{****} 

\begin{document}

\title{Face Image Reflection Removal}

\author{Renjie Wan$^{\dagger}$\\
	\and
	Boxin Shi$^{\bigstar}$\\ 
	\and
	Haoliang Li$^{\dagger}$\\
	\and	
	Ling-Yu Duan$^{\bigstar}$   \\
	\and
	Alex C. Kot$^{\dagger}$\\
	\and
	$^{\dagger}$School of Computer Science and Engineering, Nanyang Technological University, Singapore\\
	$^{\bigstar}$National Engineering Laboratory for Video Technology, School of EECS,  Peking University, China\\
}

\maketitle

\begin{abstract}
	Face images captured through the glass are usually contaminated by reflections. The non-transmitted reflections make the reflection removal more challenging than for general scenes, because important facial features are completely occluded. In this paper, we propose and solve the face image reflection removal problem. We remove non-transmitted reflections by incorporating inpainting ideas into a guided reflection removal framework and recover facial features by considering various face-specific priors. We use a newly collected face reflection image dataset to train our model and compare with state-of-the-art methods. The proposed method shows advantages in estimating reflection-free face images for improving face recognition.
\end{abstract}

\section{Introduction}
As one of the most commonly observed subjects in computer vision, face images are often captured by various types of imaging sensors under unconstrained wild scenarios, which bring different types of distortions to the clear face images. When face images are captured behind a piece of glass, the reflection-contaminated face images not only unpleasantly affect the human perception, but also degrade the performance of visual computing algorithms focusing on face. Therefore, it is of great interest to remove the reflections and enhance the visibility of the human faces behind the glass.    
\begin{figure}[t!]
	\centering
	\includegraphics[width=1.0\linewidth]{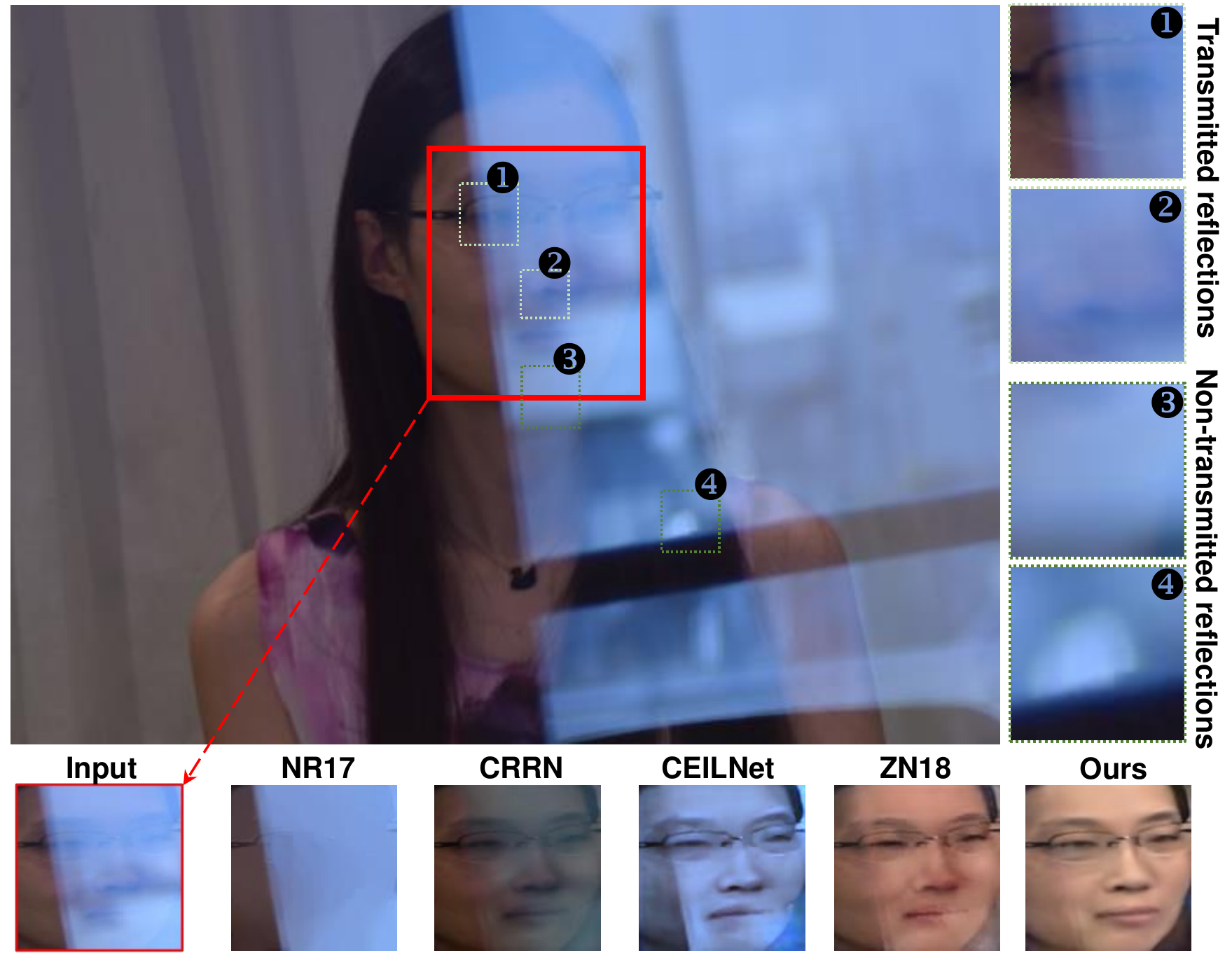}	
	\caption{Examples of transmitted reflections with high transmittance, non-transmitted reflections with low transmittance, and the reflection removal results obtained by using NR17~\cite{Nikolaos2017reflection}, CRRN~\cite{wan2018crrn}, CEILNet~\cite{fan2017generic}, ZN18~\cite{Zhang2018Single}, and our method.  
	}
	\label{fig:firstexample}
\end{figure}

Different from general objects or scenes, faces have its specific priors awarded by humans, even if a slight reflection (transmitted) distortion may significantly annoy human perception~\cite{liu2007face}. When reflection become stronger (non-transmitted), machine vision algorithms may fail due to the lost or distortion of important facial features. How to {\it remove non-transmitted reflections} and {\it recover important features for machine vision methods} pose unique challenges for face reflection removal.

Existing reflection removal methods~\cite{li2014single,wan2016depth,wan2018crrn,fan2017generic} can be directly applied to face images with reflections. However, due to ignorance of the specific facial priors and the non-transmitted reflections, artifacts on face largely remain on the recovered `reflection-free' face image (\eg, the result obtained by CEILNet~\cite{fan2017generic} in~\Fref{fig:firstexample}). Thus, methods designed for generic reflections in arbitrary scenarios are not capable to deal with these challenges. To recover the facial information largely occluded by non-transmitted reflections, it is also straight-forward to integrate specific facial priors into the single image inpainting methods (\eg,~\cite{pathak2016context,zhang2018demeshnet}). However, solely relying on learned representations from the training data to inpaint the reflection-contaminated region may not faithfully retain the lost face identity feature.

\begin{figure*}[t!]
	\centering
	\includegraphics[width=1.0\linewidth]{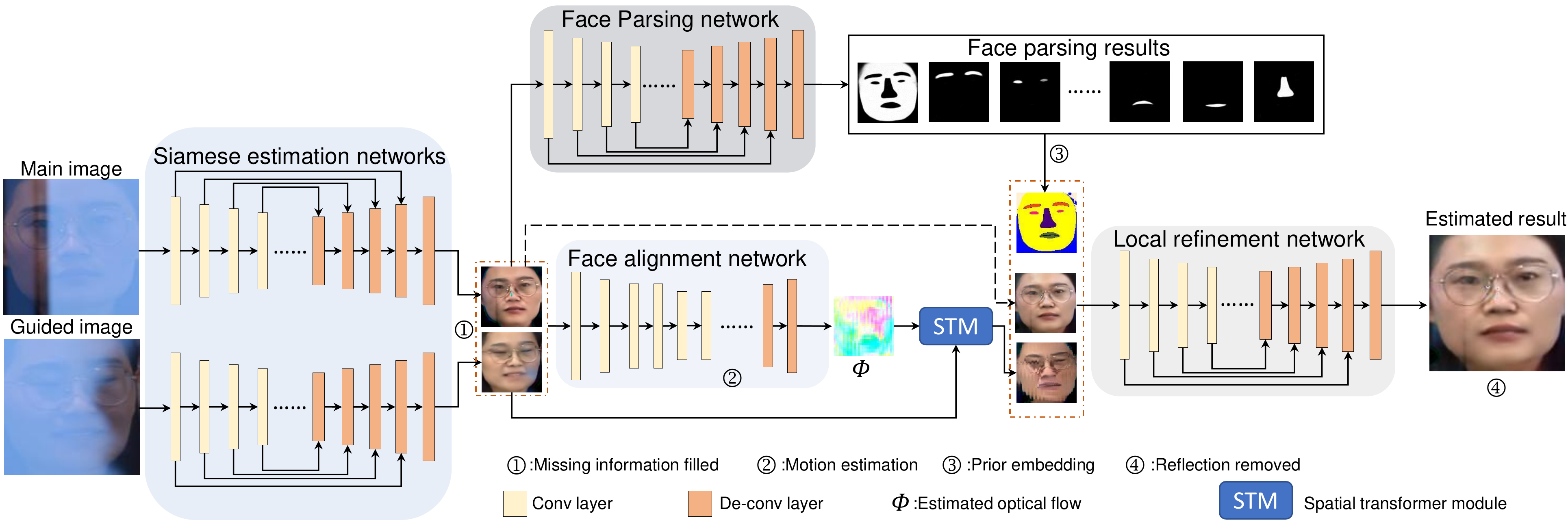}	
	\caption{The framework of our proposed network. It consists of four networks with distinctive functions: the siamese estimation networks to roughly estimate the missing facial information (\textcircled{1}), the face alignment network to compensate the different motion direction between two face images (\textcircled{2}), the face parsing network to estimate the face parsing maps and do the prior embedding (\textcircled{3}), and the local refinement networks to refine the local details (\textcircled{4}). 
	}
	\label{fig:framework}
\end{figure*}

To conquer the above challenges, we first explore the complementary
advantages from image inpainting and reflection removal to recover the facial information occluded by non-transmitted reflections. Then, to recover important facial features, we employ the guided removal framework with particular considerations on the feature level similarity and specific facial priors. Instead of only predicting the missing information from the learned representations, the guided framework~\cite{wan2017sparsity,li2018learning} can provide more accurate identity details of the human faces due to the additionally guided information, from an additional
face image of an image sequence with continuous face movement, which provide different facial pose or reflection properties. On the other hand, since the face similarity is compared in a compact feature space and the pixel level similarity adopted by previous reflection removal methods can hardly guarantee the identity consistency~\cite{zhang2018demeshnet}, we also embed the feature level similarity and other specific facial priors into the estimation process.  

Our complete framework is shown in~\Fref{fig:framework}, which includes four components: the siamese estimation network to roughly estimate the missing facial information, the face alignment network to compensate the different motions between two face images, the prior estimation network to embed the facial priors into the whole estimation process, and the local refinement network to refine the local details. Our major contributions are summarized as follows:
\begin{itemize}
	\itemsep=0.01cm
	
	\item We propose the first reflection removal framework that targets at the face images for improving human perception and facilitating machine vision algorithms.
	
	\item We propose an effective approach to remove non-transmitted reflections by mixing the merits of image inpainting and reflection removal.  
	
	
	
	\item We recover important facial features by employing a guided removal framework with particular considerations on the feature level similarity and specific facial priors.
	
	
	
	
	
	\item We build the first face reflection image dataset to facilitate the research of reflection removal in a specific domain and  accordingly perform quantitative and qualitative evaluation.

	
	
\end{itemize}

\section{Related work}
\begin{figure*}[t!]
	\centering
	\includegraphics[width=1.0\linewidth]{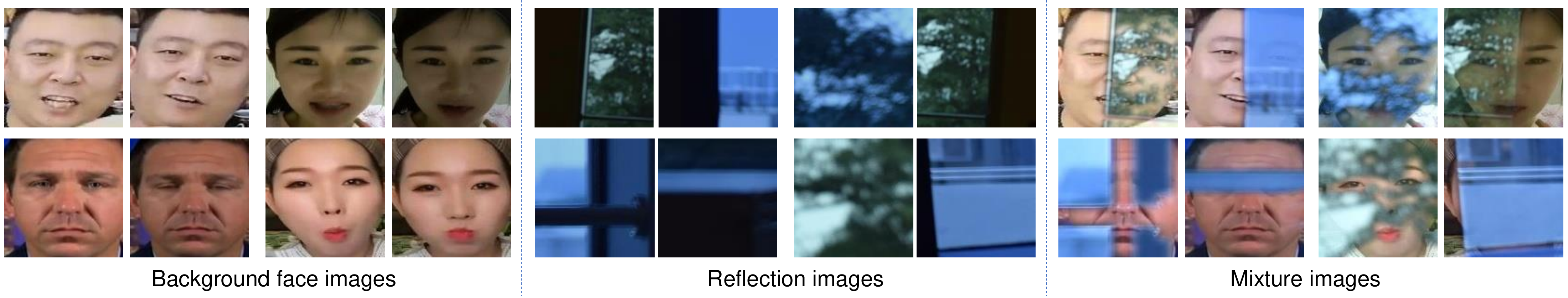}	
	\caption{The background face images, reflection images, and the mixture images in our training dataset.
	}
	\label{fig:faceimages}
\end{figure*}
\paragraph{Reflection removal.}
Previous works on reflection removal can be roughly classified into two categories. The first category solves by using the non-learning based methods. For example, Li~\etal~\cite{li2014single} and Nikolas~\etal~\cite{Nikolaos2017reflection} made use of the different blur levels of the background and reflection layers. Shih~\etal~\cite{shih2015reflection} used the GMM patch prior to remove reflections with the visible ghosting effects. The handcrafted priors adopted by these methods are based on the observations of some special properties between the background and reflection (\eg, different blur levels~\cite{wan2016depth,li2014single}) which is often violated in the general scenes especially when these properties are weakly observed. 

The deep learning framework also benefits reflection removal problems. For example, Fan~\etal~\cite{fan2017generic} proposed a two-stage deep learning approach to learn the mapping between the mixture images and the estimated clean images. Recently, Wan~\etal~\cite{wan2018crrn} also proposed a concurrent model to better preserve the background details. The method proposed by Zhang~\etal~\cite{Zhang2018Single} first utilized the generative model to better learn the mappings from the mixture image to the clean images. However, existing methods are all designed for general scenes, which have difficulty in preserving facial details in face image reflection removal problem.

\vspace{-10pt}
\paragraph{Face image enhancement.} Numerous methods have been proposed during the past decades to solve different face image enhancement problem including face hallucination~\cite{liu2007face}, face deblurring~\cite{pan2014deblurring}, and face inpainting~\cite{lin2007quality}. Recently, the end-to-end deep learning framework are introduced to solve this problem in a data-driven manner. For example, Li~\etal~\cite{li2017generative} proposed a method based on the generative model to solve the face inpainting problem. Chen~\etal~\cite{chen2018fsrnet} made full use of the geometry prior to solve the face super-resolution problem. Shen~\etal~\cite{shen2018deep} also proposed a method to solve the face deblurring problem by using the face semantic priors. However, the face image reflection removal problem has never been explicitly modeled and solved.


\section{Proposed method}
In this section, we describe the dataset, the design methodology of the proposed reflection removal network, the optimization process, and the training details.
\subsection{Dataset preparation}
The data-driven approaches need a large-scale dataset to learn the inherent reflection properties~\cite{wan2018crrn}. Previous methods~\cite{wan2018crrn,fan2017generic,Zhang2018Single} obtain the training dataset by using the following image formation model:
\begin{equation}
\I = \alpha \B + \beta \R, 
\label{eq:formationmodel}
\end{equation} 
where $\alpha$ and $\beta$ are the mixing coefficients and $\I$, $\B$, and $\R$ are the mixture image, the background image, and reflection image, respectively. The background images $\B$ can be obtained from generic image datasets~(\eg, PASCAL~\cite{Everingham10} or COCO~\cite{lin2014microsoft}) only when targeting the reflections at arbitrary scenes~\cite{wan2018crrn,fan2017generic,Zhang2018Single}. Accordingly, existing benchmark reflection removal datasets (\eg, SIR$^2$~\cite{SIR2-iccv17}) are not suitable for our ask due to the scenery diversity. Although many face image datasets have been proposed~(\eg, CELEBA~\cite{liu2015faceattributes} and CASIA Webface dataset~\cite{yi2014learning}), they are also not applicable for our problem since they mostly consider a fixed facial pose. To facilitate the training and evaluation of our approach, we build a large-scale face image training dataset collected from online resources and its corresponding evaluation dataset taken in the real world. 
\vspace{-10pt}

\paragraph{Training dataset.} Our background face images in the training dataset are collected from Youtube by cropping face images from several consecutive video frames. The reflection images are taken by ourselves based on the method proposed in~\cite{wan2018crrn}. Then, we generate the mixture images by using~\Eref{eq:formationmodel}. To focus on the vital facial components, we further adopt the MTCNN~\cite{7553523} to crop the face portion. We show samples from our training dataset in~\Fref{fig:faceimages}. 

Our training dataset has two major characteristics: 1)\textbf{Diversity.} The face images in the training dataset are with different races, expressions, and poses; 2) \textbf{Scale}: The training dataset have $15950$ face images from approximately $450$ people to meet the request of data-driven methods and each person is labeled by their corresponding person IDs.
\vspace{-10pt}

\paragraph{Evaluation dataset.} The images in the evaluation dataset are all taken in the real world by using different capturing devices (\eg, DSLR cameras and mobile phones) with diverse scene settings. Except for the face images occluded by reflections, we also take its corresponding `groundtruth' images with same identity information for further evaluation. The evaluation dataset has $450$ images from $25$ different person.
\begin{figure*}[t!]
	\centering
	\includegraphics[width=1.0\linewidth]{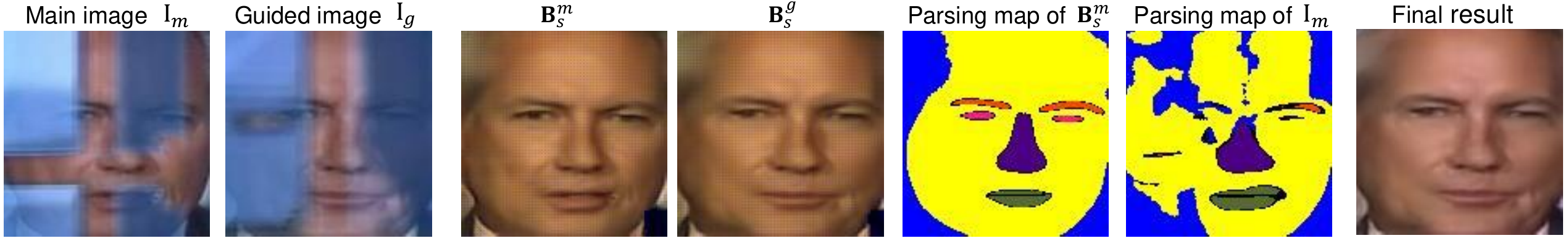}	
	\caption{The intermediate results $\B_{s}^{m}$ and $\B_{s}^{g}$ obtained by the siamese estimation network. The parsing maps of $\B_{s}^{m}$ and $\I_{m}$ obtained by the face parsing network and the final result of the local refinement network rightmost.
	}
	\label{fig:interresult}
\end{figure*}

\subsection{Network architecture}
As shown in~\Fref{fig:framework}, our network includes four parts to do the missing facial information estimation, motion compensation, prior estimation, and local refinement, respectively. Except that the face alignment network is largely based on an existing optical flow estimation network~\cite{fischer2015flownet}, other three networks all have a similar  mirror-like framework with the encoder to capture the context information by contracting the feature channels step by step and decoder part to obtain the final results. 

\subsubsection{Siamese estimation networks}
\label{sec:sen} 

Due to the occlusions caused by the reflections, it is non-trivial to estimate facial priors (\eg, facial landmark positions and parsing) from the reflection-contaminated input images directly. We first use the siamese estimation networks with shared weights to roughly estimate the coarse face images from the input image pair as:
\begin{equation}
\{\B_{s}^{m}, \B_{s}^{g}\} = \{\G_{s}(\I_{m}), \G_{s}(\I_{g}) \},
\end{equation}
where $\G_{s}$ denotes one branch of the siamese estimation networks, $\I_{m}$ and $\I_{g}$ are the main image and guided image with different reflections or varying facial properties (\eg, pose and illuminations), and $\B_{s}^{m}$ and $\B_{s}^{g}$ are the roughly recovered images corresponding to $\I_{m}$ and $\I_{g}$, respectively.

\paragraph{Local context loss.} Due to the regional property of reflections~\cite{wan2018region}, the missing information occluded by reflections cannot be well reconstructed by solely minimizing the global loss on the whole image. To solve this problem, we adopt the local context loss widely used by the image inpainting methods~\cite{pathak2016context} as: 
\begin{equation}
\ll_{c} = \|\W\odot(\G_{s}(z)-z^{\ast})\|_{1},
\label{eq:localcontextloss}
\end{equation}
where $z$ is the input to the network, $z^{\ast}$ is its corresponding ground truth, $\odot$ is the element-wise product operation, and $\W$ denotes the binary mask corresponding to the non-transmitted reflections. When the reflections occupy the whole image plane, \Eref{eq:localcontextloss} degrades to the common global loss.

\paragraph{Adversarial loss.} To roughly estimate the missing information occluded by the non-transmitted reflections, we employ the Conditional Wasserstein GAN as follows:
\begin{equation}
\begin{aligned}
\ll_{\mathbf{adv}} &= \min_{\G_s}\max_{D_s\in \mathcal{D}} ~E_{z,z^{\ast}\sim \P_{r}}[D_s(z, z^{\ast})] \\&- E_{z\sim \P_{r}}[D_s(z,\G_{s}(z))],
\end{aligned}
\end{equation}
where $D_S$ is the discriminator network, $\mathcal{D}$ is the set of $1$-Lipschitz functions and $\P_{r}$ is the real data distributions. Our discriminator takes an input image with a size of $128 \times 128$ and has 6 strided convolutional layers followed by the ReLU activation function. In the last layer, we use the sigmoid function to generate the final result. 

Combining the above terms, the loss function for the siamese estimation networks become:
\begin{equation}
\begin{aligned}
\ll_{\mathbf{SEN}} =& ~\alpha_{e}(\ll_{1}(\B_{s}^{m}, \B^{m}) + \ll_{c}(\B_{s}^{m}, \B^m) +\ll_{1}(\B_{s}^{g}, \B^{g})\\ & +\ll_{c}(\B_{s}^{g}, \B^g)) 
+
\lambda_{e}\ll_{\mathbf{adv}}^{m} +\beta_{e}\ll_{\mathbf{adv}}^{g},
\end{aligned}
\label{eq:sen}	
\end{equation}
where $\ll_1$ is the classical pixel-wise loss and $\alpha_{e}=0.5$, $\lambda_{e} = 10^{-4}$, and $\lambda_{e} = 10^{-4}$ are weights to balance different terms. 

The siamese estimation networks can be regarded as a {\it single-image approach} to solve this problem. From the results shown in~\Fref{fig:interresult}, using such networks alone is not sufficient to efficiently remove the reflections on face. However, from the face parsing results shown in~\Fref{fig:interresult}, it helps to alleviate the difficulties for estimating the facial priors in the next stage by estimating some key facial components. 

\subsubsection{Face parsing network}

Previous methods~\cite{li2017generative,chen2018fsrnet} have proved the effectiveness of the face specific prior knowledge in preserving the essential appearance information and rough locations of the facial components.

To better recover the important facial features, we embed the facial prior into our estimation process by employing the face parsing network.
We use the U-Net as the backbone but just keep the vital layers for the efficiency of the whole network to estimate the parsing maps of facial components as follows:
\begin{equation}
\M = \G_{p}(z),
\end{equation}
where $\G_{p}$ is the face parsing network, $z$ denotes the input images of this network, and $\M$ denotes the $11$-channel semantic parsing map of the important facial components as shown in~\Fref{fig:framework}. 

\subsubsection{Face alignment network} 
\label{sec:fan}
Due to the misalignment between the input image pair, their motion or pose inconsistency may increase the difficulty to recover image details~\cite{xue2017video}. To compensate the motion inconsistency, we adopt FlowNetS model~\cite{fischer2015flownet} as the backbone to build the face alignment network. It takes the two roughly recovered images from the siamese estimation networks as the input and aims at estimating the optical flow from $\B_{s}^{m}$ to $\B_{s}^{g}$ as shown in~\Fref{fig:framework} as follows:
\begin{equation}
\pphi = \G_{f}(\B_{s}^{m},\B_{s}^{g}),
\end{equation}
where $\G_{f}$ denotes the face alignment network and $\pphi$ denotes the estimated optical flow field. With the flow field $\pphi$, the guided image can be warped to the main image by using the spatial transformer network~\cite{jaderberg2015spatial} as:
\begin{equation}
\begin{aligned}
&\B^{w}_{s}(i,j) =\\& \sum_{h,w\in \mathcal{N}} \B^{g}_{s}(h,w) {\rm M}(0, 1-|\pphi_{ij}^{y}-h|) {\rm M}(0,1-|\pphi_{ij}^{x}-w|),
\end{aligned}
\end{equation}
where ${\rm M} =\max(\cdot, 0)$, $\pphi_{ij}^{x}$ and $\pphi_{ij}^{y}$ denote the predicated $x$ and $y$ coordinates for the pixel $\B^{w}_{s}(i,j)$ and $\mathcal{N}$ represents the four-pixel neighbors of $(\pphi_{i,j}^{x}, \pphi_{i,j}^{y})$. 

\paragraph{Landmark loss and face parsing loss.} The classical FlowNetS model~\cite{fischer2015flownet} is trained by using the supervised training strategy. However, due to the lack of corresponding ground truth optical flow, the supervised training strategy is not applicable in our settings. To sovle this problem, an unsupervised training strategy is proposed in~\cite{jason2016back} by minimizing the MSE loss between a warped image and another non-warped image. However, due to the roughly estimate results $\B_{s}^{m}, \B_{s}^{g}$ may have different colors, the MSE loss cannot serve as a valid way to measure the difference. Thus, we propose to use the facial landmarks and the face parsing results to facilitate the training process. 

In order to align $\B_{s}^{m}$ and $\B_{s}^{g}$, the landmarks of $\B_{s}^{m}$ and those of the warped image $\B^{w}_{s}$ should be close to each other. We first get the facial landmarks corresponding to $\B_{s}^{m}$ and $\B^{w}_{s}$ and then define the landmark loss as follows:
\begin{equation}
\ll_{lam} = \|\{\ttheta^{m}\}_{i,j=1}^{68} - \{\ttheta^{s}\}_{i,j=1}^{68} \|_{2}^{2},
\end{equation}
where $\{\ttheta^{m}\}_{i,j=1}^{68}$ and $\{\ttheta^{s}\}_{i,j=1}^{68}$ are the facial landmarks corresponding to $\B_{s}^{m}$ and $\B^{w}_{s}$, respectively. 


For the face parsing loss, we first feed  $\B_{s}^{m}$ and $\B^{w}_{s}$ to the face parsing network to get corresponding parsing maps and then the face parsing loss is defined as:
\begin{equation}
\ll_{fap} = \|\G_p(\B_{s}^{m}) - \G_p(\B^{w}_{s})\|_{2}^{2}.
\end{equation} 

By combining the above the two terms, the loss function for the face alignment network becomes:
\begin{equation}
\ll_{\mathbf{FAN}} = \ll_{lam} + \lambda_{a}\ll_{fap},
\label{eq:fan}	
\end{equation}
where $\lambda_{a} = 0.8$.

\subsubsection{Local refinement network}
\label{sec:localrefinementnetwork}
To suppress the artifacts from the output of the siamese estimation networks (the second and third images shown in~\Fref{fig:interresult}) and better preserve the facial components, the main image after the siamese estimation networks $\B_{s}^{m}$ ($3$ channels), the warped guided image ($3$ channels), and the probability maps of facial label ($11$ channels) are concatenated into a $17$ channel tensors as the input to the local refinement network as:
\begin{equation}
\B^{m} = \G_{o}([\B_{s}^{m}, \B_{s}^{w}, \G_{p}(\B_{s}^{m})]),
\end{equation}
where $\G_o$ is the local refinement network, $\B^{m}$ denotes the final estimated results, and $\G_{p}(\B_{s}^m)$ is the probability maps of facial labels corresponding to $\B_{s}^m$.

\paragraph{Statistic identity loss.} Existing reflection removal methods always aim at estimating the recovered images with higher PSNR~\cite{li2014single} and/or SSIM values~\cite{wan2018crrn} by using different pixel-wise loss functions. Though these pixel-wise loss functions are simple to calculate, the recovered face images estimated by them may have a larger difference from the ground truth since the face similarity is more properly defined in a compact feature space rather than the image pixel space~\cite{zhang2018demeshnet}. Two perceptually indistinguishable face images with high SSIM values still have quite obvious feature-level differences~\cite{zhang2018demeshnet}. 

To solve this problem, existing face restoration methods~\cite{shen2018deep,deng2017uv} adopt the perceptual loss to measure the high-level feature similarity as:
\begin{equation}
\ll_{i} = \sum_{l}||\F_{l}(z^{\star})-\F_{l}(z)||_{2}^{2},
\label{eq:oldstatistics}
\end{equation}
where $\F_{l}$ denotes the $l$-th layer features from a pre-trained loss network (\eg, VGG16~\cite{simonyan2014very}) and $z^{\star}$ and $z$ denote the estimate images and targets, respectively. 

However, \Eref{eq:oldstatistics} can only calculate the first-order statistics of the feature level differences. Previous methods~\cite{li2017second, koniusz2017higher} have shown the important roles of the higher-oder statistics in different tasks. Based on the perceptual loss used by previous methods~\cite{shen2018deep,deng2017uv} in~\Eref{eq:oldstatistics}, we propose a statistic identity loss to measure the feature level similarity on the basis of maximum mean discrepancy (MMD) in the local refinement network. As a kind of distribution divergence measurement derived from kernel embedding, MMD can measure the similarity of two distributions based on all-order moments as used in the two-sample testing problem~\cite{haoliang_2018_CVPR}. Given two images $z$ and $z^{\star}$, MMD is defined as:
\begin{equation}
\mathbf{MMD}(z, z^{\star}) = \left \| \mu_{\P}(z) - \mu_{\P}(z^{\star})] \right \|_{\H},
\end{equation}
where $\H$ denotes the Hilbert space and $\mu_{\P}$ is defines as:
\begin{equation}
\mu_{\P} := \mu(\P) = \mathbb{E}_{z \sim \P}[\phi(\cdot)] = \mathbb{E}_{z\sim \P}[k(z,\cdot)].
\end{equation}
Here, $\phi : \R^{d} \to \H$ is a feature map, and $k(\cdot,\cdot)$ is the kernel function induced by $\phi(\cdot)$. Combining these, our statistic identity loss becomes: 
\begin{equation}
\ll_{sti} = \|\frac{1}{N_{z^{\star}}}\phi(\F_{l}(z^{\star}))^{\top} \mathbf{1}_{z^{\star}} - \frac{1}{N_{z}} \phi(\F_{l}(z))^{\top} \mathbf{1}_{z}\|_{F}^{2},
\end{equation}
where $\mathbf{1}_{z^{\star}}$ and $\mathbf{1}_{z}$ are all-one vectors with the size $N_{z^{\star}}$ and $N_{z}$, respectively. $\frac{1}{N_{z^{\star}}}\phi(\F_{l}(z^{\star}))^{\top} \mathbf{1}_{z^{\star}}$ and $\frac{1}{N_{z}}\phi(\F_{l}(z))^{\top} \mathbf{1}_{z}$ are the empirical measure~\cite{gretton2007kernel} of $\mu_{\P}(z^{\ast})$ and $\mu_{\P}(z)$, respectively.

\paragraph{Local structural facial loss.} Since human vision is more sensitive to the key components (\eg, eyes, lips, and mouths)~\cite{shen2018deep}, instead of solely minimizing the global loss on the whole face image, we use the local structural facial loss similar to~\cite{shen2018deep} to better preserve the facial information as follows:
\begin{equation}
\ll_s(\B^{m}, \B^{\ast}) = \sum_{k = 1}^{K}\|M_{k}(\G_{p}(\B_{s}^{m}))(\B^{m}-\B^{\ast})\|_{1},
\end{equation}
where $M_{k}(\cdot)$ denotes the binary operation. We impose the local structural facial loss on eyebrows, eyes, noses, lips, and teeth. 

Then the loss functions for the local refinement becomes: 
\begin{equation}
\ll_{\mathbf{LRN}} = \lambda_{o}\ll_{1}(\B^m, \B^{\ast}) + \alpha_{o}\ll_{s}(\B^m, \B^{\ast}) + \beta_{o}\ll_{sti}(\B^m, \B^{\ast}),
\label{eq:lcl}
\end{equation}
where $\lambda_o = 1.5$, $\alpha_o = 0.5$, and $\beta_o = 5$.
\begin{figure*}[t!]
	\centering
	\includegraphics[width=1.0\linewidth]{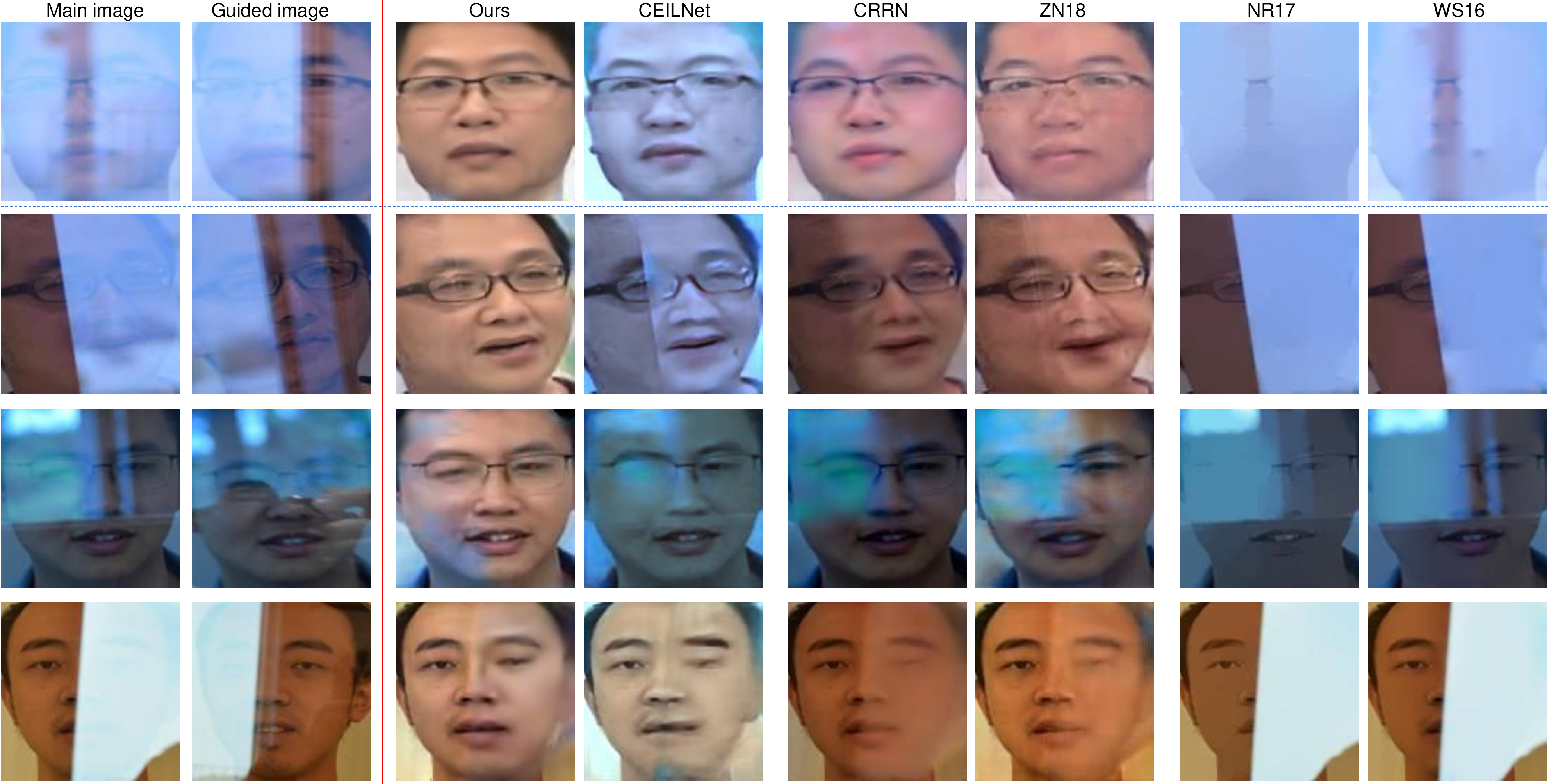}	
	\caption{Examples of reflection removal results on the evaluation dataset, compared with CEILNet~\cite{fan2017generic}, CRRN~\cite{wan2018crrn}, ZN18~\cite{Zhang2018Single}, NR17~\cite{Nikolaos2017reflection}, and WS16~\cite{wan2016depth}. More results can be found in the supplementary materials.
	}
	\label{fig:resultfaceimages}
	\vspace{-15pt}
\end{figure*}

\subsubsection{Overall loss function}
Combining $\ll_{\mathbf{SEN}}$ in \Eref{eq:sen}, $\ll_{\mathbf{FAN}}$ in \Eref{eq:fan}, and $\ll_{\mathbf{LRN}}$ in \Eref{eq:lcl}, our overall loss function for training is defined as follows:
\begin{equation}
\ll =  \ll_{\mathbf{SEN}} + \ll_{\mathbf{FAN}} +  \ll_{\mathbf{LRN}}.
\end{equation}
\subsection{Implementation and training details}
We have implemented our model using PyTorch. The complete training processes of our network can be divided into two stages: 1) We train the siamese estimation networks, the face parsing network, and the face alignment network separately to convergence. 2) We fix the face parsing network and then combine them with the local refinement network, and the entire network is fine-tuned again, which grant more opportunities to cooperate accordingly. The landmarks in~\Sref{sec:fan} are obtained by using a pretrained landmark estimation network based on the MobileNet~\cite{howard2017mobilenets}. We adop the pretrained LightCNN~\cite{wu2018light} as the face recognition model used in the statistics identity loss of~\Sref{sec:localrefinementnetwork}. The learning rate for whole network training is set to $5 \times 10^{-5}$ for the first 30 epochs and then decreases to $5 \times 10^{-6}$. 

\section{Experiments}

We first compare the visual quality and quantitative errors of our method and state-of-the-art reflection removal approaches. We also conduct a user study to investigate how each method improves human perception. Then, another experiments on face identity recognition are conducted to investigate whether our proposed method can contribute to the high-level face recognition algorithms. At last, we conduct an ablation study to verify the effectiveness of each component in our network.

\subsection{Comparison with the state-of-the-arts}
We compare our method with state-of-the-art reflection removal methods, including ZN18~\cite{Zhang2018Single}, CRRN~\cite{wan2018crrn}, CEILNet~\cite{fan2017generic}, NR17~\cite{Nikolaos2017reflection}, and WS16~\cite{wan2016depth}. For fair comparison, we use the released codes of the above methods and train all models with the same training dataset for the data-driven methods (CRRN~\cite{wan2018crrn}, CEILNet~\cite{fan2017generic}, and ZN18~\cite{Zhang2018Single}).

\paragraph{Visual quality comparison.} We first show examples of recovered reflection-free face images by our method and other five methods in~\Fref{fig:resultfaceimages} to check their visual quality. In these examples, our method removes reflections more effectively and recovers the details of the face images more clearly. All the non-learning based methods (NR17~\cite{Nikolaos2017reflection} and WS16~\cite{wan2016depth}) cannot remove the non-transmitted reflections effectively and also downgrade the visual quality of the regions not covered by reflections. Though the data-driven based methods performs much better than the non-learning based methods, the final estimated results still remain visible artifacts and some key face components are also wrongly estimated (\eg, ZN18~\cite{Zhang2018Single} in the second examples). CRRN~\cite{wan2018crrn} and CEILNet~\cite{fan2017generic} cause serious color degradation in the estimated results. It is mainly due to linear dependency between the mixture image and background image of their image formation models.
\vspace{-10pt}

\begin{figure}[t!]
	\centering
	\includegraphics[width=1.0\linewidth]{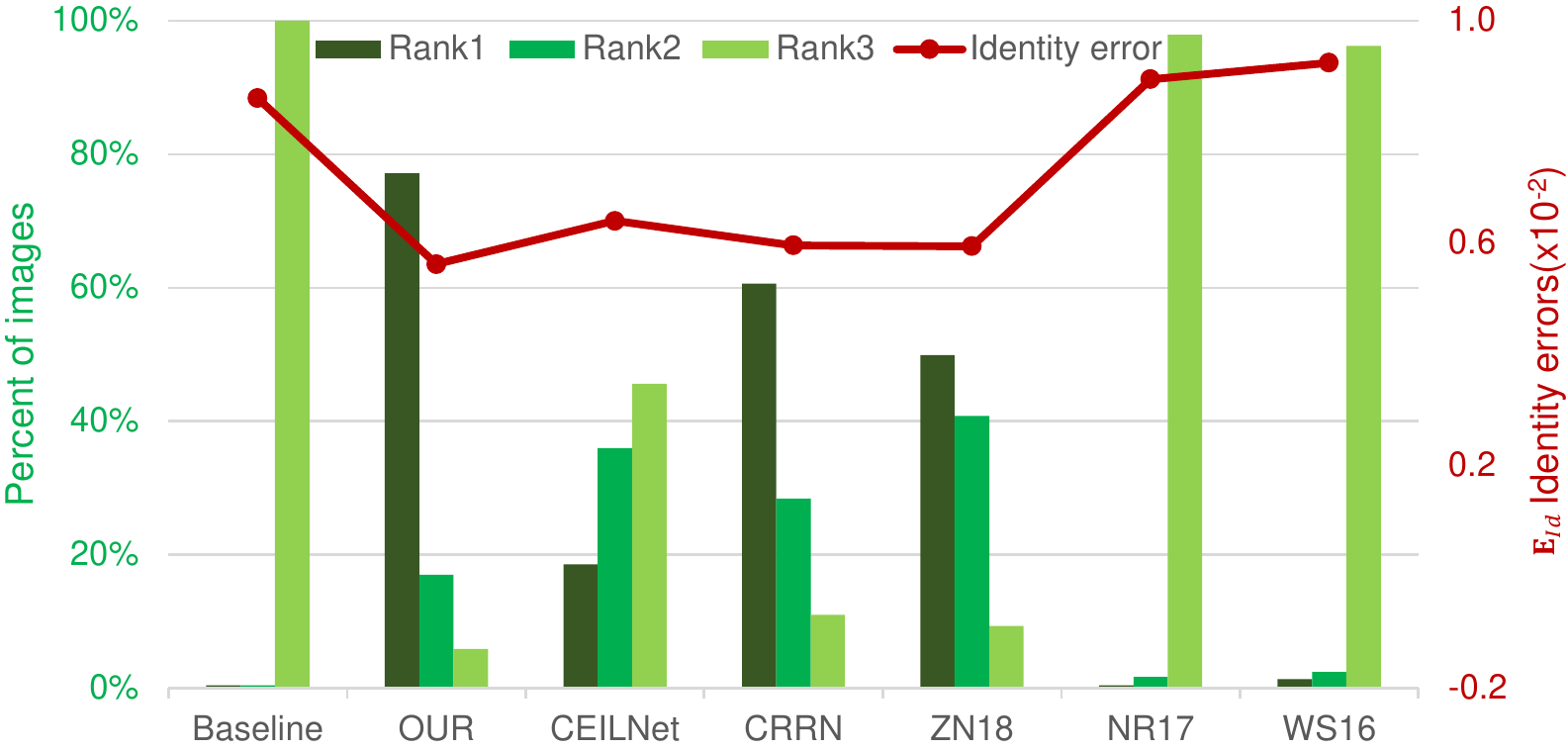}	
	\caption{The human perception study and quantitative comparisons in terms of the Identity errors on the baseline, our method, CRRN~\cite{wan2018crrn}, ZN18~\cite{Zhang2018Single}, CEILNet~\cite{fan2017generic}, NR17~\cite{Nikolaos2017reflection}, and WS16~\cite{wan2016depth}. For the human perception study, we use the input mixture image as the baseline. For the quantitative comparison, lower identity error values are better and we use the errors between the `ground truth' and the input mixture image as the baseline.
	}
	\label{fig:identitymetrics}
\end{figure}

\paragraph{Quantitative comparison.}  Since we consider the image capturing in a dynamic scenario, it is difficult to obtain the well-aligned ground truth like previous methods~\cite{SIR2-iccv17,Xue2015A}. Thus, the widely used error metrics (\eg, PSNR and SSIM) are not suitable for our evaluations due to the lack of well-aligned ground truth. Instead, we evaluate the performances from the feature domains by using the high-level facial information. We use the OpenFace toolbox~\cite{amos2016openface} to compute the identity distance between the `ground truth' face images and different results obtained by using identity error defined as: $\Ee_{\mathbf{Id}} = \| \F_{E}(\B) - \F_{E}(\B^{\ast})\|_{2}^{2}$,  where $\F_{E}$ denotes the face recognition model used in the evaluations. 

From the results shown in~\Fref{fig:identitymetrics}, our method achieves the best identity scores, which demonstrates that the proposed method preserves the face identity well. The two non-learning methods WS16~\cite{wan2016depth} and NR17~\cite{Nikolaos2017reflection} achieves even worse results than the baseline, which are consistent with the observations in the visual quality comparisons. The results of deep learning based methods are much better than that of the non-learning based methods. However, since CEILNet cannot well recover the color information, its performance also cannot beat CRRN~\cite{wan2018crrn} and ZN18~\cite{Zhang2018Single}. CRRN~\cite{wan2018crrn} and ZN18~\cite{Zhang2018Single} achieves similar performances. However, due to the wrongly recovered face components, its average scores is also worser than our proposed methods. 
\vspace{-10pt}

\paragraph{Human perception evaluations. } To investigate how each method improves human perception on reflection-removed results, we conduct another experiments based on the user study scores. We invite 30 participants to judge all images in our evaluation dataset. The participants are required to give three rankings for different results. From the results shown in~\Fref{fig:identitymetrics}, nearly $80\%$ percent of our images are given the first rank, which are best among all methods. The two non-learning based methods generally fails on almost all images. The other three learning based methods (CEILNet~\cite{fan2017generic}, CRRN~\cite{wan2018crrn}, ZN18~\cite{Zhang2018Single}) perform much better than the non-learning based methods. The ranking of how each method performs in human perception evaluation (the higher rank-1 score  the better) is generally consistent with quantitative comparison (the lower error the better).
\vspace{-10pt}

\paragraph{Face recognition evaluations.}
\begin{table}
	\small
	\centering
	\caption{Quantitative evaluation results using four different error metrics, and compared with~FY17\cite{fan2017generic}, NR17~\cite{Nikolaos2017reflection}, WS18~\cite{wan2018region}, and LB14~\cite{li2014single}.}	
	\begin{tabular}{|P{1.6cm}|P{0.9cm}|P{0.9cm}|P{0.9cm}|P{1.0cm}|P{0.9cm}|}
		\hline
		& Top-$1$ & Top-$3$  & Top-$5$ & Top-$10$ \\ \hline
		Baseline & 7.95\% & 2.5\% & 12.5\% & 15.91\% \\ \hline		
		Ours & \textbf{53.41\%} & \textbf{69.32\%}  & \textbf{76.14\%} & \textbf{85.23\%} \\ \hline
		CRRN~\cite{fan2017generic} & 52.27\% & 60.23\% & 67.05\% & 72.73\% \\ \hline
		ZN18~\cite{Nikolaos2017reflection} & 50.00\% & 65.91\% & 72.73\% & 80.68\%  \\ \hline
		CEILNet~\cite{wan2018region} & 40.91\% & 59.09\% & 68.18\% & 80.68\% \\ \hline
		NR17~\cite{li2014single} & 1.14\% & 9.09\% & 3.41\% & 3.41\% \\ \hline
		WS16~\cite{wan2016depth} & 6.82\% & 9.09\% & 1.25\% & 1.82\% \\ \hline
	\end{tabular}
	\label{fig:resulttable}
\end{table}

\begin{figure*}[htb]
	\centering
	\includegraphics[width=1.0\linewidth]{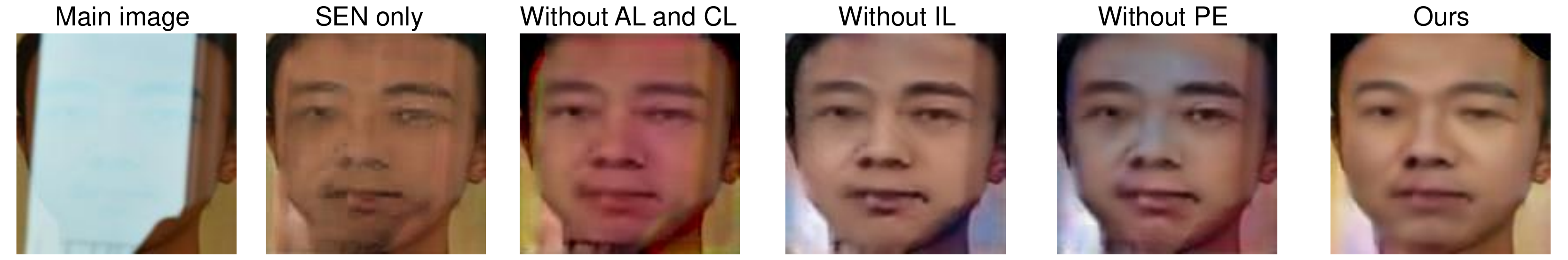}	
	\caption{Examples of our complete model against our model with only the siamese estimation network (SEN), our model without the adversarial loss (AL) and local context loss (CL), our model without prior embedding (PE), and our model without the identity loss (IL). 
	}
	\label{fig:selfcomparison}
	
\end{figure*}

\begin{figure*}[htb]
	\centering
	\includegraphics[width=1.0\linewidth]{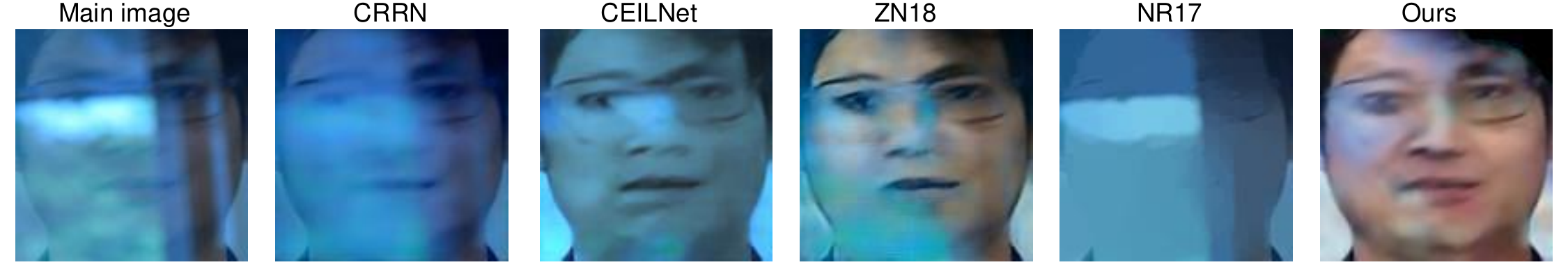}	
	\caption{Extreme examples with diverse reflections and degraded face color, compared with CRRN~\cite{wan2018crrn}, CEILNet~\cite{fan2017generic}, ZN18~\cite{Zhang2018Single}, and NR17~\cite{Nikolaos2017reflection}.
	}
	\label{fig:extreme}
\end{figure*}
The identity errors and human perception study in~\Fref{fig:identitymetrics} partly reveals the network ability of preserving the face identity information. In order to fully investigate whether our method can improve the accuracy of machine vision algorithms, we evaluate our estimated results in the task of face recognition. Given a probe face example, the goal of recognition is to find an example from the gallery set that belongs to the same identity~\cite{li2017generative}. We randomly select 575 identities from the LFW dataset~\cite{learned2016labeled} and then merge it with the identities in our evaluation dataset to form an evaluation dataset with roughly 600 identities. Each identity has roughly the same amount of images in each set. 

We use the Top-$1$, Top-$3$, Top-$5$, and Top-$10$ recognition accuracy to evaluate the performances. From the results shown in~\Tref{fig:resulttable}, our method achieves the highest recognition accuracy than all other methods. The non-learning based methods can not effectively increase the recognition rate and their results are even lower than the baseline. The learning based methods achieve much better results. However, the artifacts observed in~\Fref{fig:resultfaceimages} downgrade their performances.  ZN18~\cite{Zhang2018Single} achieves similar performances in the Top-$1$ part. However, the higher scores among other parts prove the effectiveness of our proposed method.

\subsection{Network analysis}


\begin{table}
	\small
	\centering
	\caption{Idendity errors of our complete model against our model with only the siamese estimation network (SEN only), the model without the adversarial loss and local context loss (W/o AL and CL), the model without the identity loss (W/o IL), and the model without the prior embedding (W/o PE). Lower value is better.}	
	\begin{tabular}{|P{0.6cm}|P{1.3cm}|P{2.1cm}|P{1.0cm}|P{1.0cm}|P{0.4cm}|}
		\hline
		\textbf{Ours} & SEN only  & W/o AL and CL & W/o IL & W/o PE  \\ \hline
		\textbf{0.562} & 0.592 & 0.611 & 0.598 & 0.606  \\ \hline
	\end{tabular}
	\label{fig:selfcomparisontable}
	\vspace{-15pt}	
\end{table}

Our framework consists of four parts, \ie, the siamese estimation network (SEN), the face alignment network (FAN), the face parsing network (FPN), and the local refinement network (LRN). In this section, we have conducted several experiments to further analyze the contributions of the guided reflection removal framework  and the influence of different loss functions.

The first one is to show the effectiveness of the guided reflection removal framework. We conduct this experiment by only keeping one branch of the SEN. As discussed in \Sref{sec:sen}, this setting can be regraded as the straight-forward single-image approach to solve this problem, which still contains obvious artifacts in the final estimated results as shown in~\Fref{fig:selfcomparison}. The identity loss in~\Tref{fig:selfcomparisontable} also proves this phenomenon, where it has relatively poor performance when compared with the complete model. Then, another experiment is conducted to verify the concepts leveraged from the image inpainting technique by removing the local context loss and adversarial loss. The two loss functions aim at estimating the missing information occluded by the non-transmitted reflections. From the results shown in~\Fref{fig:selfcomparison}, without the two loss functions, the network fails to estimate the facial components occluded by the reflections. 

Another experiment is to evaluate the contributions from the identity loss in the local refinement network. We train a network by removing the two loss functions. From the results shown in~\Tref{fig:selfcomparisontable} and~\Fref{fig:selfcomparison}, though the key components are successfully recovered, the color inconsistency between the regions with and without reflections are also very obvious. The lower face identity distance also prove its weakness.  Then, we remove the prior embedding mechanism in the LRN, where the input to LRN reduces to a $6$-channel tensors. From the results shown in~\Fref{fig:selfcomparison} and~\Tref{fig:selfcomparisontable}, the performances are similar to the results obtained by using the model without the identity loss. 

\section{Conclusion}
We propose and solve the face image reflection removal problem in this paper. Different from the general scenes, the special properties of face images pose challenges on non-transmitted reflection and face identity feature recover for face reflection removal. To address these issues, we first leverage ideas from image inpainting to recover the key facial components occluded by reflections, and we then utilize the guided removal framework, prior embedding and statistics identity loss to better recover the important facial features. Based on the newly collected training dataset, our framework achieves better performances than existing methods on the proposed evaluation dataset.

\paragraph{Limitations.} The performances of our method may drop when the reflections on faces become non-uniform as shown in~\Fref{fig:extreme}. However, even in this situation, our method still outperforms other methods. Another limitation of our work is from the evaluation dataset, as it is difficult to obtain the face images with different identities. Though we have tried our best to cover more realistic scenarios, the diversity of our evaluation dataset is still limited. Our current dataset is suitable for a proof-of-concept purpose, and we are working on increasing the diversity of our evaluation dataset by including more identities and more challenging scenes~(\eg, the images capture by surveillance cameras).

\end{document}